\documentclass[conference]{IEEEtran}
\usepackage[utf8]{inputenc}
\usepackage[T1]{fontenc}

\usepackage{ragged2e}
\usepackage[pdftex,colorlinks=true,linkcolor=black,urlcolor=black,citecolor=black,plainpages=false,pdfpagelabels]{hyperref}
\usepackage{eurosym}
\usepackage{url}
\usepackage{pdfpages}
\usepackage{todonotes}
\usepackage{figsize}
\usepackage{amssymb} 
\usepackage{amsmath}
\usepackage{amsthm}
\usepackage{xspace}
\usepackage{graphicx}
\usepackage{stfloats}
\usepackage{float}
\usepackage{algorithmic}
\usepackage{booktabs}
\usepackage{multirow}

\DeclareGraphicsExtensions{.pdf,.ai,.jpg,.png}
\graphicspath{{./figures/}} 

\usepackage{color, colortbl}

\usepackage{xcolor}
\definecolor{Gray}{gray}{0.9}

 \newtheorem{definition}{Definition}

\newcommand{\real}{\ensuremath{\mathbf{R} }}

\usepackage{enumitem}
\setlist[enumerate]{topsep=4pt,itemsep=4pt}

\begin{document}

\title{On a Uniform Causality Model for Industrial Automation}
\author{
\IEEEauthorblockN{
Maria Krantz\IEEEauthorrefmark{1}\IEEEauthorrefmark{8},
Alexander Windmann\IEEEauthorrefmark{1}\IEEEauthorrefmark{8},
René Heesch\IEEEauthorrefmark{1},
Lukas Moddemann\IEEEauthorrefmark{1},
Oliver Niggemann\IEEEauthorrefmark{1},
}
\IEEEauthorblockA{
\IEEEauthorrefmark{1}Institute of Automation\\
Helmut Schmidt University, Hamburg, Germany\\
Email: firstName.lastName@hsu-hh.de\\
\IEEEauthorrefmark{8}Equal contribution\\}
}

\maketitle

\begin{abstract}
The increasing complexity of Cyber-Physical Systems (CPS) makes industrial automation challenging. 
Large amounts of data recorded by sensors need to be processed to adequately perform tasks such as diagnosis in case of fault. 
A promising approach to deal with this complexity is the concept of causality.
However, most research on causality has focused on inferring causal relations between parts of an unknown system. 
Engineering uses causality in a fundamentally different way: complex systems are constructed by combining components with known, controllable behavior.
As CPS are constructed by the second approach, most data-based causality models are not suited for industrial automation.
To bridge this gap, a Uniform Causality Model for various application areas of industrial automation
is proposed, which will allow better communication and better data usage across disciplines. 
The resulting model describes the behavior of CPS mathematically and
, as the model is evaluated on the unique requirements of the application areas, it is shown that the Uniform Causality Model can work as a basis for the application of new approaches in industrial automation that focus on machine learning.

\end{abstract}

\section{Introduction} \label{intro}

Causality has been employed to explain the behavior of an unknown system, specifically in the fields of the social, medical and biological sciences. 

Causal inference \cite{pearl2010causal} starts with a system with known elements but unknown relations between them and, using statistical methods on usually large amounts of data, aims at resolving the underlying relationships between these elements.
In contrast, engineering starts with the parts of a system, which need to be assembled into a system with an intended function. 
The causal relation between these parts is generally known and the system is designed such that the causal relations between the individual parts achieve a certain functionality in the finished system. 
In a way, this is the inverse of causal inference---known causal relations between the elements are used to construct a hierarchical system with a target function. 
%
Therefore, it would be useful and necessary to have a definition of causality specific to mechanical and plant engineering. 
Specifically in the field of Cyber-Physical Systems (CPS), causal relations are not only known theoretically, but can also be measured through the sensors in the system. 
Furthermore, causalities are central to most problem solving applications in CPS, such as production planning and diagnosis, which will be discussed below. 
A definition of causality in CPS would be beneficial for 
using causality in the context of machine learning, which has been successfully applied to a multitude of engineering systems.
However, causality has only recently seen increased attention \cite{Scholkopf.2021}. 

\paragraph{Causality}
In order to be useful for industrial automation, a model of causality in a CPS has to fulfil several requirements.

Requirement 1: A mathematical formalism must be capable of representing the concept of cause and effect (i.e. causal relations) in CPS. 

Requirement 2: Likewise, we need a mathematical formalization of the concept of a component (e.g. drives, production modules) within a CPS. 

Plants are physical systems and physical systems are defined by time.
And in physics effects appear after the cause. 

Requirement 3: Any causality model must capture time and assure that all effects occur after the cause.

Causality is often represented in graphs, namely Directed Acyclic Graphs (DAGs) \cite{pearl1995causal}. 
DAGs do not allow cycles, which means they do not allow relations such as A influencing B and B influencing A. 
However, control circuits employ exactly this structure. 

Requirement 4: A causality model for CPS must be capable of representing control circuits. 

To derive further requirements, we analyzed the usage of causality in two main Artificial Intelligence (AI) application fields in industrial automation:

\paragraph{Production Planning}
In general, planning denotes the task of finding a sequence of actions leading from the initial state to the goal state. 
In the context of production systems, planning determines the way a product is manufactured \cite{rogalla_improved_2018}. 
The equivalent of CPS in the field of production systems are {Cyber-Physical Production Systems} (CPPS). 
For simplicity, we assume a fully automated, modular CPPS, whose modules can be freely combined to create a functional event chain. 
Each module offers a functionality which can be applied under consideration of certain parameters and is triggered in the form of a function call.  
Thus, production planning in the field of CPPS aims to find the sequence of functionalities as well as the corresponding parameter sets that are needed in order to manufacture the product. The sequence can be processed automatically, allowing for a faster adaptation to changing product property requirements. 

Requirement 5: It must be possible to model materials and products, including their initial and target state.


Requirement 6: Using the mathematical formalism, it must be possible to model the functionalities of modules, including the effect of potential parameters.
 

Requirement 7: The causality model must allow for an interchangeable configuration of the modules of the CPPS.

\paragraph{Diagnosis}
Diagnosis is the task of computing the root cause of a fault based on potentially incomplete observations.
While observations of the system working correctly is available abundantly, data of faulty components and their subsequent effects are rare.
In order to apply diagnosis, it is necessary to differentiate in between faulty and normal components and to represent the expected system behavior.

 


Requirement 8: Components may be OK or not-OK (KO). The resulting behavior of the components and how faults propagate through the system must be representable. 





In this paper, we derive a mathematical formalism for causality in CPS which fulfills the requirements of applications in industrial automation. 
On top of contributing to research of causality in CPS, the model advances the fields of planning and diagnosis of CPS, specifically the use of machine learning algorithms in these fields, by enabling these algorithms to use causality. 
The formalism presented here is a first step towards the development of an algorithm which can formalize and transfer causalities between systems. 


\section{State of the Art} \label{sota}


Causality is often analyzed
using the approach of causal inference. 
It uses experimental observations, often data sets from large-scale and high-throughput experiments, including interference such as mutations, to draw conclusions about the causal network. 
As causal relations cannot be directly read off from the data, they have to be inferred by settings such as randomized control trials, where the randomization counteracts confounding factors \cite{eberhardt2017introduction}. 
\cite{pearl1995theory} proposed a definition of inferred causation which is based on the idea that for causation to exist between two variables, a strictly directed path between these two variables needs to exist in a minimal model. 
\cite{pearl1995causal} also introduced the use of directed acyclic graphs for causal inference and described the do-operator, which can be used to mathematically describe an intervention in the model. 
New additions to the theory of causal inference have been made recently, expanding the tool box for analysis of complex data of unknown networks, such as the works by \cite{meinshausen2020foundations} and \cite{hyttinen2017constraint}. 
\\
  Interactions of a system with its environment can be described via Halpern and Pearl models (HP models), which model the world using a set of endogenous and exogenous variables and sets of structural equations \cite{halpern2016}. 
Furthermore, modeling approaches such as state machines encode causal knowledge and can be executed to compare the behavior of the simulated system with real-world data. 
A finite-state machine can only be in one state at a given time and can transition from one state to another in reaction to external inputs \cite{wang2019formal}.
Within the transition, a causal relation is encoded: each transition is triggered by an external factor, which causes the state machine to transition into the next state. 
Causal dependencies can be inferred from plant documentation to extract engineering information about the plant and from process data. 
An automated generation of these causal models was examined by \cite{Arroyo.2016} for causal graphs and by \cite{song2020} for functional process models. 

In industrial automation, an application closely connected to causality is the planning of a production process, which deals with the order of functionalities within the process to produce a certain product. 
A functionality performs a transformation of one or more properties of the processed product. The transformation is caused by the behavior of the physical components and fundamental laws of physics.
Since causalities are the basis for the transformations, \cite{Erdem.2012} presented a causality-based planning approach for reconfigurable factories, where possible actions are described as a set of causal laws. 
Within recent years the AI planning community developed approaches to use so called action model learning for AI planning such as \cite{garrido2020learning} and \cite{grand2022tempamlsi}. 
Action model learning approaches are based on symbolic learning, which means finding rules underlying the regarded data sets. 
The resulting action model is describing the valid transitions in the space of states of a planning environment.
\cite{Onaindia.2018} argued that symbolic learning can be exploited to determine causal relations by using the underlying action model of a domain. 
\cite{pfips} identified requirements an AI-planning approach has to meet to be applicable in the field of flexible production system and compared different approaches with regard to these requirements. 
\\
Despite the existing theoretical approaches of AI planning, there are no concrete and successful implementations in industrial automation so far, because real world planning problems are very complex and the prerequisites, such as a modular CPPS, are still being developed.

  A major challenge in time series data is the detection of abnormal situations and the location of root causes. 
  Such systems are difficult to learn, since it is not only required to capture the temporal dependency of each time series, but causal relations between different pairs of time series \cite{zhang2019}.
  Diagnosis can be performed using several approaches, which differ in the effort required to model every possible system behavior. 
  Model-based diagnosis, as described in First-Order Logic within the method of \cite{reiter1987} and \cite{dekleer1992}, focuses on the investigation of whether a detected fault signature is consistent with a set of mode signatures \cite{feldman2009}. 
Model-based diagnosis can be divided into two subcategories, namely weak-fault models \cite{dekleer1992} and strong-fault models \cite{struss1989physical}. 
Strong-fault models specify faulty behavior modes for its components.
An elaborate process is required to declare every faulty signature from all component combinations in a complex system according to its causalities.
In contrast, weak-fault models indicate a system with its normal behavior \cite{feldman2010approximate}. Therefore, the system description only needs to be formulated in the healthy state, without every single fault mode of a system.
Consistency-based diagnosis belongs to the weak fault approach. Here, the consistency of the system is checked by comparing the learned model of normal behavior against the actual behavior. Inconsistency between observable components can be detected and a diagnosis of the system can be performed on this basis. For accurate identification of the fault, a sufficient number of observable components is important. However, if there are only partial observable components \cite{buenoimproved} proposed the diagnostic coefficient that uses the likelihood of the observations as the basis of discrepancy indicator to solve this partial observability case.
\\
  

\section{Solution} \label{solution}


CPS offer a unique possibility to describe and understand causality, because causal effects in between components and subsystems are in principle well understood, and the close monitoring of the system via sensors makes these causal effects quantifiable.
Every application area in industrial automation has its own unique requirements.
However, rather than formulating a model for each task, this section proposes a Uniform Causality Model for industrial automation (UCM) that describes causality in a CPS in a way that is useful for a broad range of application areas. 

\subsection{A Uniform Causality Model}

Mathematically, every sensor and actuator in a CPS can be described as a random variable, i.e. a measurable function $X:\Omega \to \real$ from a probability space $(\Omega, \mathcal{F}, \mathbb{P})$ to $\real$. 
CPS are dynamic systems, thus the model describing a sensor can be expanded to a stochastic process: $Y^{}:\Omega \times \real \to \real, (\omega,t)\mapsto Y_t^{}(\omega)$.
For every $t \in \real$, $Y_t^{}$ defines a random variable with distribution $P_{Y_t}=\mathbb{P}\circ Y_t^{-1}$.
For example, this distribution could be a standard normal distribution, which would be denoted as $Y_t \sim \mathcal{N}(0, 1)$.
In general, the distribution underlying the sensor values does not change by itself over time.
However, it might shift due to an intervention, which motivates the first definition of a causal effect:

\begin{definition}[Effect]
\label{def:effect}
An \textit{effect} is observed when the distribution of a stochastic process $Y$ shifts from $t_1$ to $t_2$, i.e. $P_{Y_{t_1}}\neq P_{Y_{t_2}}$.
\end{definition}

In a CPS, the sensor value at a given point in time can generally be described with only a few distributions that depend on the system's underlying state. 
The sensor then transitions in between these distributions, often performing predictable cycles.

\begin{definition}[State]
\label{def:state}
Let $Y$ denote a stochastic process that models a sensor. 
Suppose there exists a finite set of distributions $S_Y$, such that for all $t \in \real$, there exists $P_Y \in S_Y$ such that $Y_t \sim P_Y$.
Such a set $S_Y$ is defined as the set of states of the sensor $Y$.
\end{definition}

The set of all observable sensors $\mathcal{Y}=\{Y^{(1)},\dots,Y^{(n)}\}$ and their states $S=S_{Y^{(1)}} \times \dots \times S_{Y^{(n)}}=\prod_{i=1}^n S_{Y^{(i)}}$ describe the underlying physical system.
A component or subsystem $\Sigma$ of a CPS can be described by the subset of the sensors $\mathcal{Y}_\Sigma \subsetneq \mathcal{Y}$ that measure its properties and their states $S_\Sigma=\prod_{i\in N_\Sigma} S_{Y^{(i)}}$, where $N_\Sigma \subsetneq\{1,\dots,n\}$ denotes the indices of the subsystems' sensors. 
However, a subsystem is more than the sum of its observable characteristics.
What defines a subsystem---and what is actually the reason why it is installed in a CPS in the first place---is its behavior in the system, i.e. how it transforms properties of other objects and components. 
Therefore, another way to think about a subsystem is as a function: $f_\Sigma:S_\Sigma \to S$. 

\begin{definition}[Subsystem]
\label{def:subsystem}
Let $\mathcal{Y}=\{Y^{(1)},\dots,Y^{(n)}\}$ denote the set of stochastic processes that model all sensors of a CPS with associated states $S=\prod_{i=1}^n S_{Y^{(i)}}$. 
A \textit{subsystem} $\Sigma=(\mathcal{Y}_\Sigma, S_\Sigma, f_\Sigma)$ of a CPS can be described with a subset of sensors that measure its properties $\mathcal{Y}_\Sigma\subsetneq \mathcal{Y}$, the set of states of the subsystem $S_\Sigma=\prod_{i\in N_\Sigma} S_{Y^{(i)}}$ given the subsystems' sensor indices $N_\Sigma \subsetneq\{1,\dots,n\}$ and a function $f_\Sigma :S_\Sigma \to S$ that describes its interaction with the observable environment.
\end{definition}

The functional representation $f_\Sigma$ of the subsystem describes its behavior. 
More precisely, it describes the expected state of the system given the subsystem's current state.
By changing the subsystem's properties, the predicted state of the system changes, which might necessitate a distribution shift in other sensors.
As the function $f_\Sigma$ is deterministic, the behavior of the subsystem $\Sigma$ is reproducible, provided the subsystem does not malfunction and there is no unobservable outside influence. 
Note that while the functional representation of causality is deterministic, the sensors are modeled as stochastic processes.
The UCM thus separates causality, which always follows the same patterns, from the observable sensor values that include uncertainty.

\begin{definition}[Cause]
\label{def:cause}
Let $\Sigma=(\mathcal{Y}_\Sigma, S_\Sigma, f_\Sigma)$ denote a subsystem. If a forced distribution shift of $Y^{(i)}\in \mathcal{Y}_\Sigma \subset \mathcal{Y}$ at $t_0$ always leads to the same effect on $Y^{(j)}\in \mathcal{Y}\backslash \{Y^{(i)}\}$ at $t_1$ with $t_1>t_0$, then $Y^{(i)}$ \textit{causes} the effect on $Y^{(j)}$. 
This reproducible effect is encoded in the functional representation of the subsystem $f_\Sigma :S_\Sigma \to S$.
\end{definition}

Definition \ref{def:cause} ensures that the cause precedes the effect, which is a sensible requirement in the context of CPS. 
Furthermore, note that the distribution shift of the first stochastic process is forced, i.e. a known internal or external factor influenced the distribution shift.
A simple correlation of the stochastic processes does not guarantee a causal relation, as there might be confounders or the correlation is completely random. 
Following the notion of \cite{Pearl:2009}, there has to be an intervention to confirm a cause.



The effects of multiple subsystems can be chained.
Where to separate such a causal chain and define a subsystem is not well-defined. 
Humans typically differentiate in between subsystems based on their function. 
For example, a component in a machine might be considered to be a subsystem, provided its properties are observable.
Multiple components form a module and multiple modules eventually form the complete CPS.
This notion of causal levels works in both ways: every component consists of even smaller components down to the (sub-)atomic level. 

\begin{definition}[Causal Construction]
\label{def:causal_construction}
Two subsystems $\Sigma_1$ and $\Sigma_2$ can form a superordinate system $\Theta=(\mathcal{Y}_\Theta, S_\Theta, f_\Theta)$, with sensors $\mathcal{Y}_\Theta= \mathcal{Y}_{\Sigma_1} \cup \mathcal{Y}_{\Sigma_2}$, states $S_\Theta=S_{\Sigma_1} \times S_{\Sigma_2}$ and a functional representation $f_\Theta: S_\Theta \to S$. 
\end{definition}

The functional representation $f_\Theta$ can be obtained by sequentially applying $f_{\Sigma_1}$ and $f_{\Sigma_2}$, resulting in a causal chain with effects that develop over time.
Causally impossible combinations of states can be resolved by giving priority to one subsystem at a time.
The two subsystems might even work adversely, shifting the other subsystem's state over and over. 
The resulting system would show a cyclic behavior, similar to control circuits.



 
\subsection{Planning with the UCM}

Planning in industrial automation deals with the optimal order of the functionalities of modules to manufacture a specific product in a CPPS.
Usually, there are several raw materials that are combined and transformed in order to get the final product and unavoidable waste.
For simplicity, all of these different materials are referred to as products.
Similarly to all subsystems of the CPPS, each product has properties which can be measured via sensors.
Furthermore, at every point in time the products are expected to have specific properties, which can be modeled via a set of states.
Finally, all products interact with the environment, albeit their influence on the modules of the CPPS is limited.
Therefore, products can be fully explained as a regular subsystem of the CPPS as described in Definition \ref{def:subsystem}.

Based on the concept of underlying products, the CPPS can be divided into subsystems in a way that enables planning.
First, the products form a subsystem, which makes it possible to monitor their properties over time.
Second, components of the CPPS are grouped into modules based on their effect on the products.
Each of these modules performs a distinct transformation of certain properties of the products, which can be referred to as the functionality.
As mentioned earlier, a CPPS in the application area of planning is modular by nature, thus the causal relations in between individual modules are limited to the effects on the products. 
Such a separation of the CPPS into modules and products greatly simplifies the causal construction described in Definition \ref{def:causal_construction}.
By chaining multiple functional representations of modules, different functional chains of the system can be simulated. 
%
The result is a mathematical formalization of the functional event chain that is modular, data-driven and accounts for uncertainty.
This enables the application of machine learning algorithms, both to learn the functional representation of the modules and their optimal configuration, which would greatly enhance the automation of the whole process.

\subsection{Diagnosis with the UCM}


Diagnosis is triggered when unexpected sensor values occur and the behavior of the system changes.
However, detecting an irregular value does not directly indicate faulty behavior, as every measurement includes uncertainty.
Real faulty behavior has to be statistically significant, which means that the underlying distribution must have changed.

\begin{definition}[Anomalous Sensor]
\label{def:anomalous_sensor}
Let $Y$ describe a sensor with states $S_Y$. If at some point in time $t \in \real$ there does not exist a distribution $P_Y \in S_Y$ such that $Y_t \sim P_Y$, then the sensor is anomalous.
\end{definition}

Whether the sensors' distribution matches one of the states can be checked with a statistical test.
Thus, subtle changes to the sensor values can be detected early if they occur very often, which enables a quicker response.
Alternatively, using predefined thresholds can act as a heuristic.
%
As discussed earlier, sensor values generally do not change their distribution on their own: in order for an effect to occur, there has to be a cause.
However, there might be a faulty component in the system that causes the effect, therefore it is sensible to control all subsystems that have a direct causal relation to the anomalous sensor.

\begin{definition}[Faulty Component]
\label{def:faulty_component}
Let the subsystem $C=(\mathcal{Y}_C,S_C,f_C)$ denote a component. 
If the empirical behavior of the component $\hat{f}_C :S_C\to S$ differs from $f_C$, then the component $C$ is called faulty.
Furthermore, if the component has an effect on a sensor which thereby becomes anomalous, the component is faulty as well.
\end{definition}

By distinguishing between anomalous sensors and faulty components, a more precise root cause analysis is possible.
A sensor of a component might be anomalous despite the component itself working correctly if there is another faulty component that causes the effect.
An anomalous sensor in an otherwise normally behaving system might indicate that the sensor itself is faulty. 
The goal of diagnosis is to identify the cause of the unexpected observation, which means to identify the faulty component (per Definition \ref{def:faulty_component}) which caused the unexpected observation. For this, the knowledge of the components which exist within the CPS at hand and their causal relations according to Definition \ref{def:cause} are used. Following the causal relations between components, the cause of the faulty behavior can be identified. 


\section{Evaluation} \label{evaluation}

In order to demonstrate the usage of the causality model, the process of hardening a knife will serve as an example. 
The characteristic 'hardened' can be achieved by heating and rapid cooling of the knife, as the physical reaction of the metal to the temperature change causes hardening. In course of the manufacturing process an oven is used to transfer energy to the knife in order to heat it. We assume that the oven has a damper and is gas fired. It is possible to control the burner inside the furnace with a set value between zero and 100 to achieve the desired temperature regardless of the outside temperature. This set value has to be passed to the oven when it's functionality is called. Heating the air around the knife inside the oven causes a heat transfer. Fundamental laws of physics are describing how this causes a temperature change of the knife depending on different factors, such as the properties of the knife. \\
The dependencies in this process could be described in the causality model. This enables AI planning methods to determine the set values automatically as part of of the planning step, if the factors, which are influencing the process, such as the outside temperature, are known. Without the model it would only be possible to find out that the module oven is needed in order to harden the knife. The set value would have to be determined manually by an engineer. \\
For diagnosis, the model could help with identifying the root cause of a fault. For example, in case the inside of the oven has a temperature sensor, deviations from the expected temperature could be found. Here it is important to use time dependent expectations for the temperature within the oven, because the oven might not be expected to run at the same temperature all the time. In the example given here, the oven is colder when the knife is placed inside the oven, then the lid is closed and the oven heats up. Therefore, the expected temperature within the oven is time dependent. When the expected temperature inside the oven is not reached at the expected time, an error occurred within the system. To identify the root cause, the causal relations between the parts of the system need to be backtracked until the cause of the deviation is found. Identification of root causes can also be simplified by the causality model when certain sensors are missing. In our example, this would be the case when, for example, the knife is not hardened properly, but no internal temperature sensor in the oven exists. In case the knife is not hardened properly after the procedure, it can be concluded that it did not get heated sufficiently. The root cause for this could be a lid in the oven, which did not close properly. 

In the Introduction, we defined eight requirements that the UCM needs to fulfil. 
As a first requirement, we asked that the UCM needs to capture the concept of cause and effect within a CPS. We introduced a mathematical definition of effect in Definition \ref{def:effect} and  defined cause in Definition \ref{def:cause}. Both definitions take the fact into account that CPS are monitored by sensors and any definition of processes within the CPS needs to be based on these sensor values. Together with Definition \ref{def:state} (state), cause and effect within a CPS can be described. 
%
Requirement 2 stated that the UCM needs to be able to mathematically represent a component. We defined the concept of a subsystem in Definition \ref{def:subsystem}, which is an adequate description of a component. 
Based on the observation that CPS are physical systems, we formulated Requirement 3, which demands that the UCM must capture time. 
Our definition of a cause (Definition \ref{def:cause}) uses time to define when a forced distribution shift leads to an effect in another sensor. 
Furthermore, by explicitly using time here, we enable the representation of control circuits as described in Requirement 4.
To formulate the next requirements, we analyzed the demands in the field of planning. Requirement 5 concerned itself with the features the UCM must provide to represent materials and products. We found that the product can be described as a subsystem of the CPS. 
%
Requirement 6 stated that the UCM must be able to model the functionalities of models and how parameters influence these functionalities. A module is a subsystem of the CPS and, as such, is described by a certain set of sensors (Definition \ref{def:subsystem}). Parameters, which alter the functionality of the module, can be described as long as there are sensors measuring them. 
%
Requirement 7 deals with the interchangeability of modules within a CPPS.
When the input and output of a subsystem are described by sensors, their causal relations can be described and causal chains of subsystems can be created.
As long as sensors are available which describe the input and output of the modules, the order of these modules within the UCM can be changed.
For the last requirement, we looked at applications from the field of diagnosis. In diagnosis, it is necessary to represent the state of a component, which can be OK or not-OK. We presented a definition for the state of a sensor (Definition \ref{def:state}) and, furthermore, a definition for anomalous sensors (Definition \ref{def:anomalous_sensor}) and faulty components (Definition \ref{def:faulty_component}). Thus, it is not only possible to define a component as OK or not-OK, but also to differentiate between faulty components and faulty sensors. 
%


\section{Discussion} \label{discussion}

Despite the success of machine learning in many domains, the impact on applications on CPS has been limited.
A core issue might be that most applications do not learn the underlying causal structure of the system \cite{Scholkopf.2021}.
However, most existing approaches to learn causality are not ideally suited for CPS, because they do not leverage knowledge about the modular and hierarchical structure of CPS and often fail to model fundamental behavior such as control circuits.
To be applicable to real-world data of CPS, the UCM has to reduce the complexity of the data significantly, which is done by focusing on a few underlying states and by grouping sensors into subsystems.

The UCM has several implications for industrial automation.
First, when working with sensor data the location of a sensor in a network of modules should be made available. Without this context information, the interpretation of sensor signals is complicated and can yield incorrect results. 
By grouping sensors in subsystems, machine learning algorithms can learn the behavior of individual modules and their interaction. 
With a concept of causal relation between components, automatic planning can arrange components into causal modules based on its effect on a product.
When the behavior of these modules is modeled, predictions about the system's behavior under changing configurations can be made.
Diagnosis algorithms should utilize information about the component's configuration in order to more quickly detect the root causes of faults. 
When the normal behavior of individual modules is learned and monitored, an anomaly can be detected via statistical tests early on. 
Once the behavior of a faulty component is learned and the causal relations within the system are understood, a prediction of how faults spread through the subsequent system is made possible.

In this paper, we present a first step towards the development of a formalism for causality in industrial automation, specifically for analyzing, planning and diagnosing CPS and CPPS. We term this formalism the Uniform Causality Model. 
%
We believe that the UCM is a valuable first step into developing a uniform approach to causality in all disciplines of automation.

\raggedright
\bibliographystyle{abbrv}
\bibliography{bib}

\begin{thebibliography}{10}

\bibitem{Arroyo.2016}
E.~Arroyo, M.~Hoernicke, P.~Rodr{\'i}guez, and A.~Fay.
\newblock Automatic derivation of qualitative plant simulation models from
  legacy piping and instrumentation diagrams.
\newblock {\em Computers {\&} Chemical Engineering}, 92:112--132, 2016.

\bibitem{dekleer1992}
J.~De~Kleer, A.~K. Mackworth, and R.~Reiter.
\newblock Characterizing diagnoses and systems.
\newblock {\em Artificial intelligence}, 56(2-3):197--222, 1992.

\bibitem{buenoimproved}
M.~L. de~Paula~Bueno, A.~Hommersom, and P.~Lucas.
\newblock An improved diagnostic method for probabilistic consistency-based
  diagnosis.
\newblock {\em Kalpa Publications in Computing}, 4:65--77, 2018.

\bibitem{eberhardt2017introduction}
F.~Eberhardt.
\newblock Introduction to the foundations of causal discovery.
\newblock {\em International Journal of Data Science and Analytics},
  3(2):81--91, 2017.

\bibitem{Erdem.2012}
E.~Erdem, K.~Haspalamutgil, V.~Patoglu, and T.~Uras.
\newblock Causality-based planning and diagnostic reasoning for cognitive
  factories.
\newblock In {\em Proceedings of 2012 IEEE 17th International Conference on
  Emerging Technologies {\&} Factory Automation (ETFA 2012)}, pages 1--8. IEEE,
  2012.

\bibitem{feldman2010approximate}
A.~Feldman, G.~Provan, and A.~Van~Gemund.
\newblock Approximate model-based diagnosis using greedy stochastic search.
\newblock {\em Journal of Artificial Intelligence Research}, 38:371--413, 2010.

\bibitem{feldman2009}
A.~Feldman, G.~M. Provan, and A.~J. van Gemund.
\newblock Solving strong-fault diagnostic models by model relaxation.
\newblock In {\em IJCAI}, volume~9, pages 785--790. Citeseer, 2009.

\bibitem{garrido2020learning}
A.~Garrido and S.~Jim{\'e}nez.
\newblock Learning temporal action models via constraint programming.
\newblock In {\em ECAI 2020}, pages 2362--2369. IOS Press, 2020.

\bibitem{grand2022tempamlsi}
M.~Grand, D.~Pellier, and H.~Fiorino.
\newblock Tempamlsi: Temporal action model learning based on strips
  translation.
\newblock In {\em Proceedings of the International Conference on Automated
  Planning and Scheduling}, volume~32, pages 597--605, 2022.

\bibitem{halpern2016}
J.~Y. Halpern.
\newblock {\em Actual Causality}.
\newblock The MIT Press, 2016.

\bibitem{hyttinen2017constraint}
A.~Hyttinen, S.~Plis, M.~J{\"a}rvisalo, F.~Eberhardt, and D.~Danks.
\newblock A constraint optimization approach to causal discovery from
  subsampled time series data.
\newblock {\em International Journal of Approximate Reasoning}, 90:208--225,
  2017.

\bibitem{pfips}
A.~Köcher, R.~Heesch, N.~Widulle, A.~Nordhausen, J.~Putzke, A.~Windmann, and
  O.~Niggemann.
\newblock {A Research Agenda for AI Planning in the Field of Flexible
  Production Systems}.
\newblock In {\em IEEE International Conference on Industrial Cyber-Physical
  Systems (ICPS)}, 2022.

\bibitem{meinshausen2020foundations}
N.~Meinshausen, J.~Peters, T.~S. Richardson, and B.~Sch{\"o}lkopf.
\newblock Foundations and new horizons for causal inference.
\newblock {\em Oberwolfach Reports}, 16(2):1499--1571, 2020.

\bibitem{Onaindia.2018}
E.~Onaindia, D.~Aineto, and S.~Jim{\'e}nez.
\newblock A common framework for learning causality.
\newblock {\em Progress in Artificial Intelligence}, 7(4):351--357, 2018.

\bibitem{pearl1995causal}
J.~Pearl.
\newblock Causal diagrams for empirical research.
\newblock {\em Biometrika}, 82(4):669--688, 1995.

\bibitem{Pearl:2009}
J.~Pearl.
\newblock {\em Causality: Models, Reasoning and Inference}.
\newblock Cambridge University Press, New York, NY, USA, 2nd edition, 2009.

\bibitem{pearl2010causal}
J.~Pearl.
\newblock Causal inference.
\newblock {\em Causality: objectives and assessment}, pages 39--58, 2010.

\bibitem{pearl1995theory}
J.~Pearl and T.~S. Verma.
\newblock A theory of inferred causation.
\newblock In {\em Studies in Logic and the Foundations of Mathematics}, volume
  134, pages 789--811. Elsevier, 1995.

\bibitem{reiter1987}
R.~Reiter.
\newblock A theory of diagnosis from first principles.
\newblock {\em Artificial intelligence}, 32(1):57--95, 1987.

\bibitem{rogalla_improved_2018}
A.~Rogalla, A.~Fay, and O.~Niggemann.
\newblock Improved domain modeling for realistic automated planning and
  scheduling in discrete manufacturing.
\newblock In {\em 2018 {IEEE} 23rd International Conference on Emerging
  Technologies and Factory Automation ({ETFA})}, volume~1, pages 464--471,
  2018.

\bibitem{Scholkopf.2021}
B.~Schölkopf, F.~Locatello, S.~Bauer, N.~R. Ke, N.~Kalchbrenner, A.~Goyal, and
  Y.~Bengio.
\newblock Toward causal representation learning.
\newblock {\em Proceedings of the IEEE}, 109(5):612--634, 2021.

\bibitem{song2020}
M.~Song and M.~Lind.
\newblock Towards automated generation of function models from p\&ids.
\newblock In {\em 2020 25th IEEE International Conference on Emerging
  Technologies and Factory Automation (ETFA)}, volume~1, pages 1081--1084.
  IEEE, 2020.

\bibitem{struss1989physical}
P.~Struss and O.~Dressler.
\newblock "physical negation" integrating fault models into the general
  diagnostic engine.
\newblock In {\em IJCAI}, volume~89, pages 1318--1323, 1989.

\bibitem{wang2019formal}
J.~Wang and W.~Tepfenhart.
\newblock {\em Formal Methods in Computer Science}.
\newblock Chapman and Hall/CRC, 2019.

\bibitem{zhang2019}
C.~Zhang, D.~Song, Y.~Chen, X.~Feng, C.~Lumezanu, W.~Cheng, J.~Ni, B.~Zong,
  H.~Chen, and N.~V. Chawla.
\newblock A deep neural network for unsupervised anomaly detection and
  diagnosis in multivariate time series data.
\newblock In {\em Proceedings of the AAAI conference on artificial
  intelligence}, volume~33, pages 1409--1416, 2019.

\end{thebibliography}

\end{document}